\documentclass[conference]{IEEEtran}
\IEEEoverridecommandlockouts
\usepackage{cite}
\usepackage{booktabs}
\usepackage{amsmath,amssymb,amsfonts}
\usepackage{algorithmic}
\usepackage{graphicx}
\usepackage[hidelinks]{hyperref}
\usepackage{textcomp}
\usepackage{float}
\usepackage{xcolor}
\usepackage{caption} 
\def\BibTeX{{\rm B\kern-.05em{\sc i\kern-.025em b}\kern-.08em
    T\kern-.1667em\lower.7ex\hbox{E}\kern-.125emX}}
\begin{document}

\title{HealthGAT: Node Classifications in Electronic Health Records using Graph Attention Networks\\

}

\author{
\IEEEauthorblockN{Fahmida Liza Piya\textsuperscript{*}}
\IEEEauthorblockA{University of Delaware \\
lizapiya@udel.edu}
\and
\IEEEauthorblockN{Mehak Gupta\textsuperscript{*}}
\IEEEauthorblockA{Southern Methodist University \\
mehakg@smu.edu}
\and
\IEEEauthorblockN{Rahmatollah Beheshti}
\IEEEauthorblockA{University of Delaware\\
rbi@udel.edu}

}

\maketitle
\begingroup\renewcommand\thefootnote{*}
\footnotetext{Equal contribution.}
\endgroup
\renewcommand\thefootnote{1}

\begin{abstract}
While electronic health records (EHRs) are widely used across various applications in healthcare, most applications use the EHRs in their raw (tabular) format. Relying on raw or simple data pre-processing can greatly limit the performance or even applicability of downstream tasks using EHRs. To address this challenge, we present HealthGAT, a novel graph attention network framework that utilizes a hierarchical approach to generate embeddings from EHR, surpassing traditional graph-based methods. Our model iteratively refines the embeddings for medical codes, resulting in improved EHR data analysis. We also introduce customized EHR-centric auxiliary pre-training tasks to leverage the rich medical knowledge embedded within the data. This approach provides a comprehensive analysis of complex medical relationships and offers significant advancement over standard data representation techniques. HealthGAT has demonstrated its effectiveness in various healthcare scenarios through comprehensive evaluations against established methodologies. Specifically, our model shows outstanding performance in node classification and downstream tasks such as predicting readmissions and diagnosis classifications\footnote{Our code is available at \url{https://github.com/healthylaife/HealthGAT}. }. 
\end{abstract}

\section{Introduction}
The amount of data stored in electronic health records (EHRs) has grown significantly, now including an immense quantity of patient interactions and various medical information. The intricate and interwoven structure of this wealth of information presents considerable obstacles to its efficient utilization of such data and poses challenges in data integration, interoperability, and extraction of clinically relevant patterns~\cite{golmaei2021deepnote}. Graph neural Networks (GNNs) represent a class of deep neural network models adept at handling graph-structured data~\cite{4700287}. These GNN-based deep learning algorithms present a promising framework for deriving significant insights from the graphical structures of EHRs. Specifically, GNNs are used to group similar patients, which aids in creating structures that resemble networks. Subsequently, this structured data is analyzed using GNNs to identify pertinent labels, such as the appropriate medical procedures.

Previous works have demonstrated that GNNs can effectively model the diagnosis distribution in patient populations given the different combinations of morbidities, allowing for the phenotyping of similar patients~\cite{jg2022graph, wanyan2021deep}. GNNs extract features utilizing the underlying data structure which allows for automatic feature extraction from raw inputs instead of depending on manually constructed features. These features have shown improved performance in a variety of domains ~\cite{liu2022introduction}. Several graph analysis tasks, such as node classification, have extensively used GNNs~\cite{atwood2016diffusion,li2018deeper}. Because of diverse application scenarios, node classification is one of the most popular research areas in graph analysis. The specific goal of the node classification task is to use the graph information to predict a class label for each unlabeled node in the network~\cite{kazienko2012label}. To illustrate the complexity and richness of the EHR data that our GNN models aim to encapsulate, Table \ref{table:medical_services} demonstrates a nominal patient journey typically recorded in clinical settings. 

\begin{figure}[t]
  \centering 
  \includegraphics[width=0.6\columnwidth]{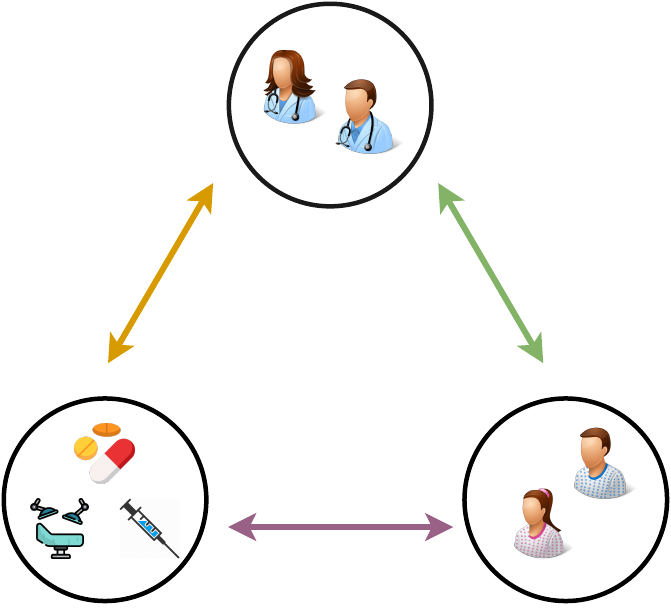} 
  \caption{Relations between different medical entities. The visual depicts three distinct entities representing doctors, patients, and medical services, including diagnoses and procedures. Doctors provide expertise and diagnoses, patients receive care, and medical services encompass various procedures and treatments. The interconnectedness of these elements illustrates the complexity and richness of EHR data, providing valuable insights for healthcare analysis and decision-making.}
  \label{fig:triangle} 
\end{figure}

Our approach involves employing GNNs to create generalized embeddings for the entities within EHR data. Some of the entities in EHRs include diagnoses, demographics, lab tests, medications, procedures, vital signs, clinical notes, and imaging reports. 
Initially, we generate embeddings for the key entities, including diagnoses, medications, and procedures. They are essential for efficient communication and data sharing among clinicians, contributing significantly to the diagnostic process. These entities in EHRs facilitate comprehensive patient care and decision-making in medical practice. The accurate recording and sharing of this information directly impacts patient outcomes and the overall effectiveness of healthcare delivery~\cite{quinn2019electronic}. Fig. \ref{fig:triangle} provides a visual representation of the relationships between different medical entities. For every patient's medical journey, each entry describes an individual encounter with a healthcare provider; these encounters are arranged according to the day of the encounter, the attending physician, and the particular medical service provided. This thorough illustration demonstrates the complex relationships found in Electronic Health Records (EHRs), providing important information for decision-making and analysis in the healthcare sector.

\begin{table}[H]
\centering
\caption{An example of a patient medical journey where `dx' and `px' represent diagnosis and medical procedure. The table below showcases a subset of EHR data, which depicts patient journeys over time.}
\label{table:medical_services}
\begin{tabular}{lll}
\toprule
Day & Doctor & Medical Service \\
\midrule
1 & A & dx-Hypertension \\
12 & A & px-MRI Scan \\
12 & B & px-Blood Test \\
20 & A & dx-Hyperlipidemia \\
20 & B & px-Prescription \\
35 & A & px-Physical Therapy \\
35 & A & dx-Type 2 Diabetes \\
35 & B & px-Insulin Therapy \\
50 & A & dx-Cardiomyopathy \\
50 & A & px-Echocardiogram \\
\bottomrule
\end{tabular}
\end{table}

Enhancing the representation of EHR data aims to improve predictions in downstream tasks. In addition to the GNN architecture, our model benefits from incorporating multiple auxiliary tasks using large-scale datasets. Auxiliary task refers to a secondary objective that aids the primary goal of the GNN model. Implementing these tasks in a multi-task learning framework has been explored in other studies, demonstrating that such an approach can enhance the primary task's performance by improving the robustness and generalizability of the model~\cite{manessi2021graph}.  It is often challenging to optimize prediction tasks, particularly for less common medical diagnoses (such as in intensive care units) or rare conditions, due to the insufficient data volume available in most health systems~\cite{NEURIPS2018_934b5358}. These auxiliary tasks provide additional information and context, augmenting the learning process and improving the overall predictive performance. 

In our study, we initiate by generating medical service embeddings, which encapsulate information about diagnoses and medical procedures performed during patient visits. These embeddings provide insights into healthcare patterns and patient care dynamics. To create these embeddings, we employ node2vec random walks~\cite{grover2016node2vec}, which capture the unique characteristics and relationships among medical services within electronic health records (EHR) data. Subsequently, we derive patient visit embeddings using a graph attention network~\cite{velickovic2017graph}, which is trained with two auxiliary tasks. By leveraging the knowledge gained from these auxiliary tasks, our model enhances node classification accuracy, leading to more reliable and accurate predictions.

The goal of our proposed work is to improve the accuracy and effectiveness of node classification tasks. Our approach involves utilizing GNNs, specifically, graph attention networks (GAT), to represent dependencies and capture complex interactions in graph structures associated with EHRs. GATs are ideal for health outcome prediction tasks, due to their innate capacity to learn from graph topology. Our primary contribution is introducing a novel GNN model tailored for node classification and downstream tasks (for instance, diagnosis prediction or readmission for a patient). We demonstrate the effectiveness of capturing the complex graphical relationships in EHRs using several experiments. Utilizing the graph structure of medical data, our model exploits the rich connections between diseases, symptoms, treatments, and patient journeys. This enables a holistic representation of disease characteristics and facilitates more accurate predictions. More broadly, the main contributions of our study are:


\begin{itemize}
\item We provide a hierarchical method for creating embeddings from EHR data using GNNs. With this approach, we create embeddings for medical codes first and then use those embeddings to improve visit embedding learning through a progressive learning mechanism.

\item We collect temporal and structural information by grouping patient admissions into visit embeddings as part of our methodology. We extract embeddings for each visit by dividing it into 24-hour segments, which integrates both structural and temporal information. 

\item We incorporate two auxiliary tasks to update visit embeddings, further improving our model. The first task relates to predicting medical codes for a particular 24-hour segment based on its embedding. This idea is expanded upon in the second task, which involves predicting the medical codes for the next 24-hour period. Using this method enables our model to extract contextual embeddings that not only reflect the medical codes in each segment but also predict the progression of the medical issue over the next 24 hours.

\item To demonstrate the effectiveness of our method, we conduct a comprehensive evaluation by comparing it against various established approaches. These experiments encompass various prediction tasks and utilize a publicly available dataset, eICU~\cite{pollard2018eicu}.


\end{itemize}

\section{Related Work}
EHRs offer a wealth of interconnected data, encompassing patient histories, diagnoses, treatments, and more \cite{guptamimic2022, GuptaMedRxiv}. On the side, GNNs are powerful tools for analyzing complex health data by mapping intricate relationships among its elements, and helping gain better insights and improve decision-making in healthcare~\cite {sun2020disease}. By employing graph representation learning, GNNs transform medical codes and their relationships into a structured network of nodes and edges. This technique effectively captures the dynamic interplay between various health components, shedding light on critical relationships such as causal or associative links between diagnoses and treatments, as well as connections between patients and healthcare providers~\cite{landi2020deep}. In EHRs, GNNs adeptly handle diverse data types and bring structure to multifaceted information, leading to the generation of meaningful patient or visit embeddings. This not only facilitates data representation but also provides valuable insights for healthcare professionals, aiding in informed decision-making and revealing underlying patterns in EHR data. The demonstrated efficacy of GNNs in enhancing patient diagnosis in previous studies highlights their potential in healthcare analytics~\cite{li2020graph, poulain2024graph}.

In our current research, we propose a unique model that harnesses the capabilities of GNNs (GNN) and focuses predominantly on medical entities frequently found in EHR data. This approach differs markedly from other established methodologies for processing generic graphs, such as \texttt{DeepWalk}~\cite{perozzi2014deepwalk}, \texttt{GraRep}~\cite{cao2015grarep}, \texttt{node2vec} and \texttt{HOPE}~\cite{ou2016asymmetric}, which are primarily unsupervised techniques aimed at preserving network structures.

While these existing methods have their strengths, they also face certain limitations. For instance, \texttt{DeepWalk} effectively generates latent representations from graph inputs, but is limited in its ability to capture higher-order graph structures and semantic information due to its reliance on random walks. This constraint can be particularly relevant in medical applications where capturing subtle relationships and temporal dynamics is crucial for accurate predictions. \texttt{LINE}~\cite{tang2015line}, on the other hand, is notable for preserving both first-order and second-order proximities, resulting in better embeddings. However, \texttt{LINE}'s dependence on doctor embeddings can be a constraint since doctor-specialty data is not always available in EHR data. Supervised graph embedding methods like \texttt{GraphSAGE}~\cite{hamilton2017inductive} and Graph Attention Network (GAT)~\cite{velickovic2017graph} also maintain graph structures, incorporating additional node or edge attributes. However, their generic graph development might not fully adapt to medical graph applications, particularly in handling the temporal sequences found in EHR data. This limitation is noteworthy as every medical service in EHR data is timestamped, presenting a significant challenge in the application of these models to the EHR domain.

To overcome these challenges and better represent diverse node types in EHRs, algorithms such as \texttt{MARU}~\cite{jiang2020maru}, \texttt{metapath2vec}~\cite{dong2017metapath2vec}, and \texttt{HNE}~\cite{chang2015heterogeneous} have been proposed. Nonetheless, their rigidity in representing different types of edges can limit their applicability in real-world healthcare scenarios. An innovative approach is proposed in~\cite{wu2021leveraging}, where a GNN is used to hierarchically learn embeddings for medical entities, doctors, and patients. This model utilizes \texttt{node2vec} for service embeddings and the \texttt{LINE} graph to derive patient embeddings from doctor and service embeddings. Despite its strengths, this method's reliance on doctor embeddings poses limitations, as such data is not typically found in EHRs.

Another significant contribution in the field focuses on scalable graph learning through random walks on continuous-time dynamic graphs~\cite{talati2021deep}. This technique is particularly effective for managing large-scale data and capturing intricate structural connections between nodes. The random walk methodology, also employed in~\cite{lu2023disease}, demonstrates the utility of this approach in medical applications like disease analysis. For random walks, word2vec~\cite{mikolov2013distributed} is applied to learn the embeddings from the generated node sequences, thus retaining structural and topological information as hidden features.

The MTGNN framework represents another significant stride in GNNs, aiming to achieve precise node embeddings for measuring disease similarity~\cite{gao2022mtgnn}. It addresses the challenge of limited labeled data for similar disease pairs through a multi-task optimization strategy, blending disease similarity tasks with link prediction tasks. This strategy is particularly effective for rare diseases, where data scarcity and complex symptom-diagnosis relationships prevail. Furthermore, the GRAM model~\cite{choi2017gram} introduces a novel approach by combining medical ontologies with a graph attention mechanism, tackling the challenges of data insufficiency and enhancing interpretability in healthcare analytics. This model exemplifies how integrating GNN techniques with medical ontologies can lead to more accurate disease predictions and a deeper understanding of medical concepts.
In contrast to the common trend followed in existing studies, our model emphasizes using commonly available medical entities within EHR data, offering a more pragmatic solution. 

In our proposed method, we also underscore the importance of utilizing auxiliary tasks, which has proven beneficial in previous studies~\cite{wu2021enhancing, lv2023semi}. For instance, \texttt{Mime}~\cite{NEURIPS2018_934b5358} uses auxiliary tasks to predict diagnoses and medications, while \texttt{node2vec}~\cite{grover2016node2vec} employs doctor's specialty prediction as an auxiliary task for learning doctor embeddings. 


\section{Methodology}
In this section, we provide an introduction to GNNs, with a focus on GATs, which serve as the foundation of our HealthGAT model. GNNs enable iterative updates of node embeddings by aggregating information from neighboring nodes using attention mechanisms. This methodology allows our model to effectively capture the nuanced relationships present within EHRs.

\subsection{\textbf{Preliminaries}}

\subsubsection{GNN}
GNNs integral to the architecture of our proposed HealthGAT model. In our GAT implementation, node embeddings are updated iteratively by aggregating information from neighboring nodes, leveraging attention mechanisms. This process enables the model to focus on relevant information from each neighbor. The aggregation in our GAT model can be represented as:
\begin{equation}
m_i = \sum_{j \in \mathcal{N}(v_i)}{attention\_weight}(v_i, v_j) \times h_j
\end{equation}

\noindent where, $\mathcal{N}(v_i)$ represents the set of neighboring nodes connected to node $v_i$ in the graph structure. \( h_j \) denotes the feature vector of a neighboring node \( v_j \) and \( \text{attention\_weight}(v_i, v_j) \) calculates the attention score between nodes \( v_i \) and \( v_j \). This mechanism allows HealthGAT to capture the interactions within EHR.

After aggregating messages using the attention mechanism, each node's embedding is updated to incorporate this collective information into its existing representation. The message, weighted by attention scores, conveys information from neighboring nodes and is used to update the embeddings of the input node. This message passing process specifically involves exchanging information between the embeddings of healthcare providers and various healthcare services to update the embeddings of the healthcare providers.

Let $\text{update}(\cdot)$ represent the function responsible for updating the node embeddings in the HealthGAT model. This function adjusts the node embeddings iteratively based on the aggregated messages obtained from neighboring nodes through the attention mechanism.

The update equation is formulated as:
\begin{equation}
h_i^{(t+1)} = {update}({message\_aggregation}(m_i), h_i^{(t)}).
\end{equation}

In this expression, $h_i^{(t)}$ denotes the embedding of node $v_i$ at iteration $t$. The $\text{update}(\cdot)$ function is designed to modify the node embedding by incorporating the aggregated messages. This iterative process is repeated for a predetermined number of cycles or until convergence. This iteration allows the HealthGAT model to capture information from various scales and levels within the graph-structured data. After completing these iterations, the final node embeddings are utilized for downstream applications such as node classification, graph classification, or link prediction, effectively reflecting the intricate relationships within the EHR data.

\subsubsection{Graph Attention Network in HealthGAT}
The HealthGAT model leverages a GAT mechanism to refine the embeddings of EHR data. In a GAT architecture, the attention mechanism plays a crucial role in message aggregation, allowing the model to focus on the most relevant information from neighboring nodes. This process can be represented as:

\begin{equation}
e_{ij} = {Attention}({W}_a \cdot h_i, {W}_a \cdot h_j)
\end{equation}

In this formulation, \( e_{ij} \) is the attention score between nodes \( v_i \) and \( v_j \), and \( \text{W}_a \) are the learnable parameters. The attention scores are normalized using a softmax function, ensuring a proper probabilistic distribution over the neighbors:

\begin{equation}
m_i = \sum_{j \in \mathcal{N}(v_i)}{softmax}(e_{ij}) \cdot h_j.
\end{equation}

This aggregation process in our model enables it to capture context-specific information and enhance the interpretability of the embeddings. Within HealthGAT, the attention mechanism allows for a more nuanced understanding of patient data, thereby providing an opportunity to better inform clinical decision-making.

\subsection{\textbf{Service Embedding}}

To facilitate service embedding, we employ the methodology introduced in \texttt{node2Vec}, which is an extension of the Word2Vec model for learning embeddings from graph-structured data. In our approach, similar to how word embeddings represent word co-occurrence frequencies in text, we estimate the co-occurrence frequency of medical services based on patient journeys. This reflects the relationship between services in a patient's healthcare journey, with infrequent co-occurrence indicating greater distance in the embedding space. The service co-occurrence frequency can be calculated as:

\begin{equation}
A_{svc(i,j)} = \sum_k {count}(s_i, s_j, \Delta t_k),
\label{eq5}
\end{equation}

\noindent where
$A_{\text{svc}(i,j)}$ is the element of the adjacency matrix representing the co-occurrence frequency between services $s_i$ and $s_j$. Furthermore, 
${count}(s_i, s_j, \Delta t_k)$ counts the occurrences of the unique pair of services $s_i$ and $s_j$ within a time window $\Delta t_k$. And, $k$ represents different time windows in the patient journeys, allowing to analyze service co-occurrence frequencies across various time intervals.

We use these embeddings to determine the importance of services in assessing patient similarity. We construct a graph of medical services, where the edges represent the co-occurrence frequency between services. To build this adjacency matrix, we analyze patient journeys within a fixed time window, ensuring it is flexible and avoids service clustering— the grouping of medical services based on similarities or co-occurrence patterns, which may distort relationships between services. Patient journeys are depicted as a sequence of medical occurrences, including diagnoses, hospitalization durations, and other healthcare interactions. This approach provides generalized knowledge about the time intervals between services, improving the transferability of the learned service embeddings.

\begin{figure}
  \centering 
  \includegraphics[width=1\columnwidth, height= 7 cm]{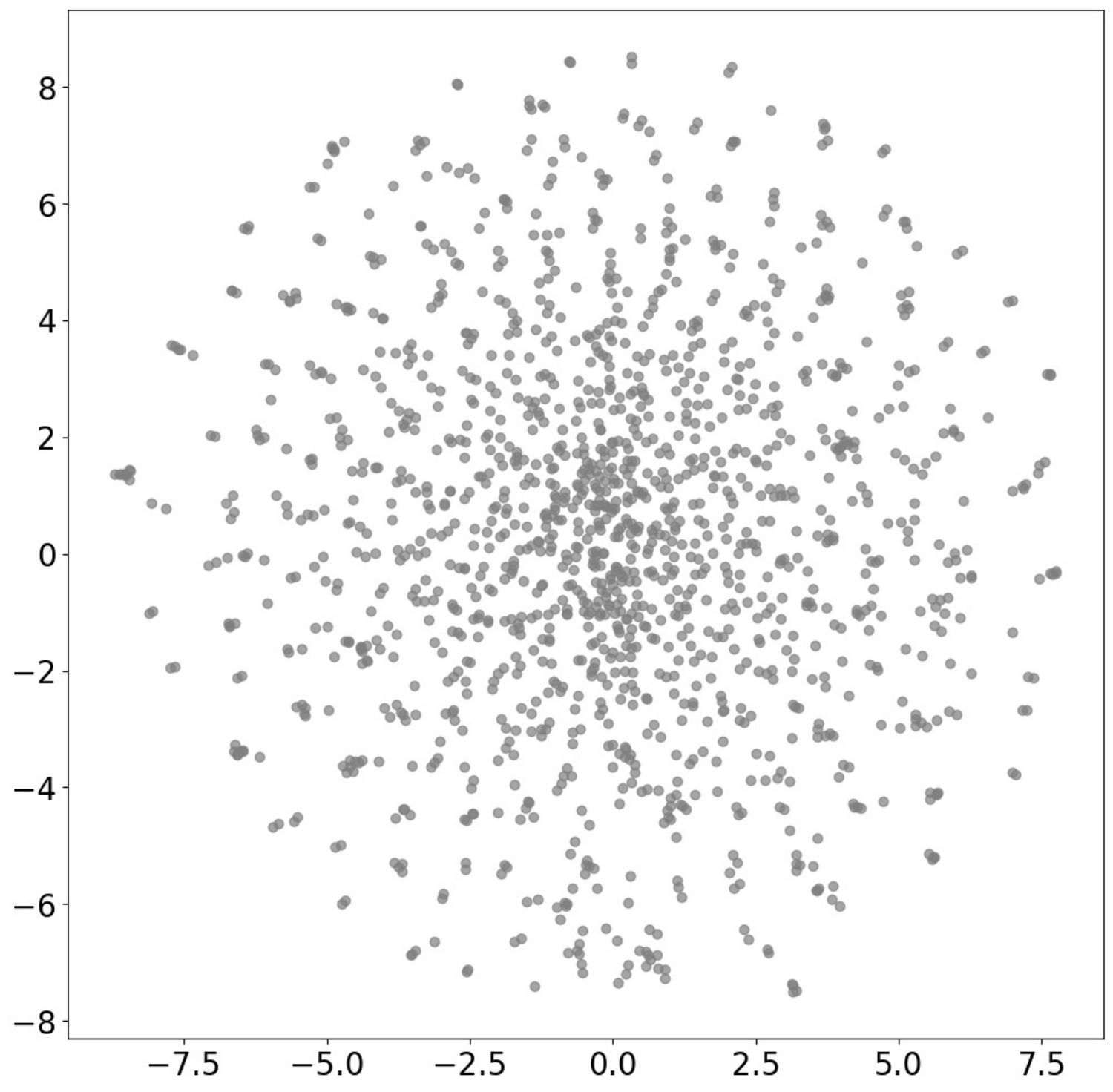} 
  \caption{Service embeddings visualized in two dimensions, with each dot representing a service. The co-occurrence and interaction patterns and distribution of services are depicted in the visualization. Services that frequently occur together or share similar contexts appear closer in the embedding space, revealing underlying structures and relationships in healthcare service provision.}
  \label{fig: visualizations of all services} 
\end{figure} 

To preserve the temporal distances between medical services, we opt for a biased random walk based embedding scheme. This approach accurately estimates a service's location in the graph by generating ``pseudo sequences", emphasizing services with higher degrees. Fig. \ref{fig: visualizations of all services} showcases the two-dimensional service embeddings derived from the dataset. Each point in this scatter plot corresponds to a unique medical service positioned according to its learned embedding. The proximity of points indicates the frequency and closeness of service co-occurrence, offering a qualitative insight into the interconnectedness of services within patient care pathways.

\begin{figure*}[hbtp]
  \centering 
\includegraphics[width=2\columnwidth]{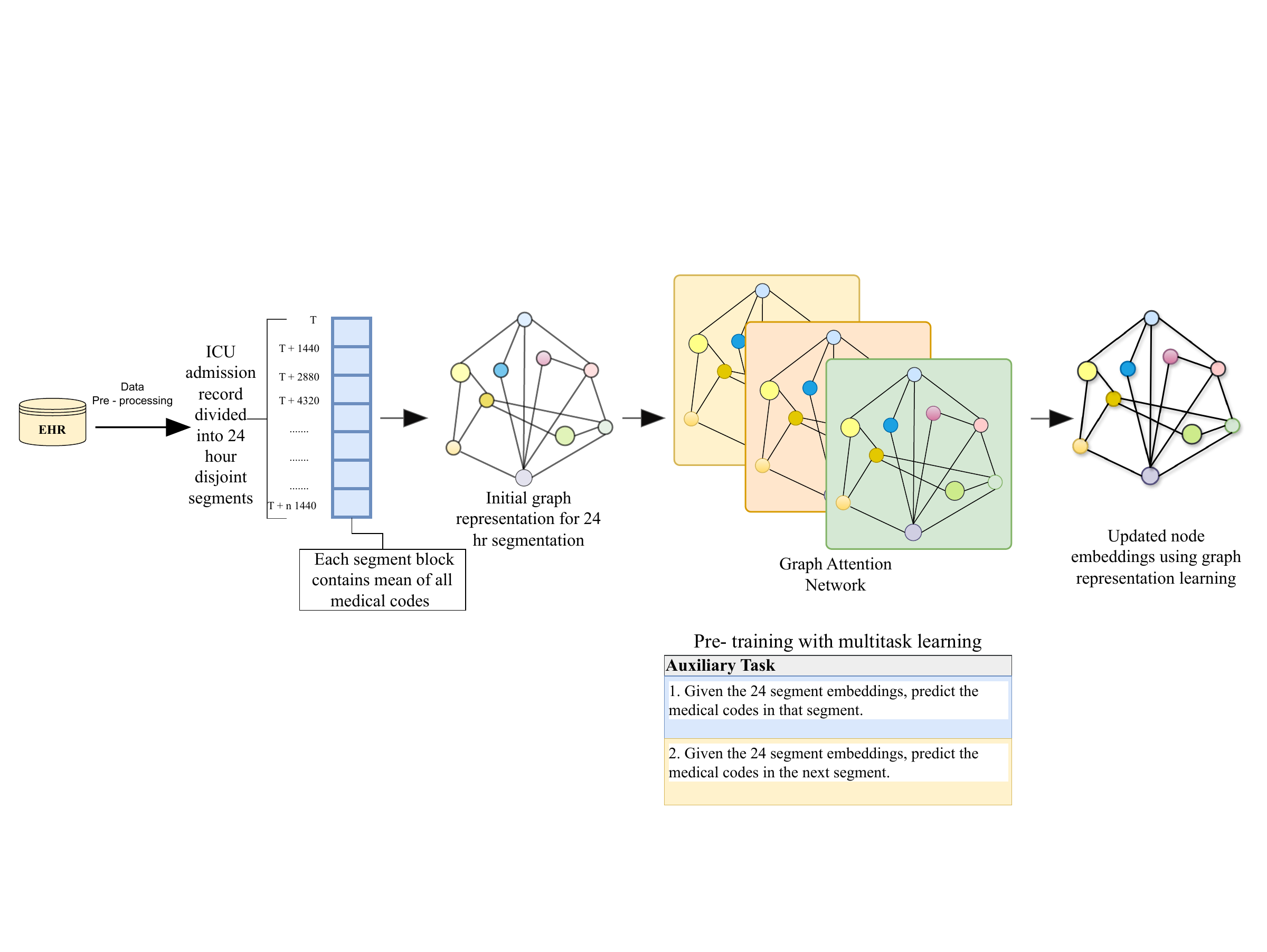} 
  \caption{The step-by-step process outlines our model's creation and refinement of visit embeddings, emphasizing the key phases involved in capturing the complex temporal dynamics of patient visits. T represents the starting point of the time intervals, corresponding to the onset of patient stays. Subsequent intervals denoted as T+1440, T+2880, and so forth, are spaced 1440 minutes (24 hours) apart from one another. The parameter n signifies the number of intervals or segments that have passed since the starting point T. Therefore, T+n1440 denotes the time point that is n intervals (each spanning 1440 minutes) ahead of T.}
  \label{fig: HealthGAT_Method} 
\end{figure*}

\subsection{\textbf{Visit Embedding}}
Understanding the trajectory of patients within healthcare systems is crucial for informed decision-making and effective healthcare delivery. This study presents a methodological framework for deriving visit embeddings, which capture the complex temporal relationships between medical services through a graph-based representation.

We divide patient stays into 24-hour intervals, where each time interval is a segment consisting of 1440 minutes, leveraging the temporal nature of patient stays. The dataset, comprising patient tables and medical journey records, is segmented to capture temporal transitions and medical interactions within these segments.

Within each segment, we compute the mean of all medical codes, utilizing previously learned medical embeddings to initialize the embeddings for every 24-hour segment within the complete admission record. The mean provides a balanced representation of all medical codes occurring within a segment, capturing the central tendency of medical interactions.

Following this, the embeddings representing the 24-hour segments undergo refinement through a GAT. We introduce two auxiliary tasks that are crucial in refining our embeddings:

\paragraph{Predicting Current Medical Codes} This task involves forecasting the medical codes in the current 24-hour segment and testing the model's ability to accurately represent the immediate medical context of a patient's stay.

\paragraph{Forecasting Future Medical Codes} Here, the model predicts the expected medical codes for the subsequent 24-hour segment. This forward-looking task assesses the model's capacity to anticipate future medical events based on current data.

These auxiliary tasks enable our model to derive contextual embeddings that not only portray the medical codes within each 24-hour segment but also capture the evolving nature of disease progression in the subsequent 24 hours.

After segmenting the 24-hour period, we compute the mean across all segments to form the initial visit embedding for each admission. We then employ a GAT again, introducing a new auxiliary task where the visit segment embedding predicts the medical codes observed throughout that visit. This iterative process refines the visit embedding, incorporating a comprehensive understanding of medical code sequences within the entire admission. This approach enriches our model's representation of the entire patient visit trajectory, offering a detailed and dynamic view of patient healthcare interactions. This process of initial embedding and subsequent refinement is depicted in Figure \ref{fig: HealthGAT_Method}. 

\section{\textbf{Experiments and Results}}
We conducted a comprehensive analysis of the HealthGAT. We picked a case study focused on node classification and readmission predictions, with a specific focus on cardiovascular diseases. We designed a series of experiments to evaluate key aspects of the model's performance.

\subsection{Dataset and Downstream Tasks}
We use the the eICU dataset~\cite{pollard2018eicu}, an extensive repository comprising EHRs from over 139,000 patients admitted to one of 335 units at 208 hospitals in the United States between 2014 and 2015. This dataset, encompassing data from more than 200,000 patients, serves as a valuable resource for critical care and healthcare analytics research. Table \ref{tab:diagnosis_categories} categorizes the major medical diagnoses within the dataset, with the 'Cardiovascular Diseases' category notably representing a substantial portion. Given its prevalence, we specifically focus on this cohort to provide insights into patient care and outcomes, leveraging it as a representative sample due to its significant presence, comprising approximately 26.04\% of the overall dataset.

\begin{table}[H]
\centering
\caption{Percentage Distribution of Patient Diagnosis Categories}
\label{tab:diagnosis_categories}
\begin{tabular}{lc}
\toprule
Category              & Percentage (\%) \\ 
\midrule
Cardiovascular        & 26.04\%         \\
Pulmonary             & 17.48\%         \\
Neurologic            & 12.14\%         \\
Renal                 & 11.23\%         \\
Gastrointestinal      & 8.94\%          \\
Endocrine             & 6.76\%          \\
Infectious Diseases   & 5.78\%          \\
Other                 & 11.63\%         \\ 
\bottomrule
\end{tabular}
\end{table}

In our study, we have identified and performed a series of downstream tasks using the eICU dataset to evaluate the effectiveness of our proposed methodology.
We perform similar classification and readmission prediction tasks as in Me2vec \cite{wu2021leveraging} and compare the results of our proposed methodology with their experiments. 
The downstream tasks include:

\textbf{Node Classification:} We classify patient records into various medical diagnosis categories. This task involves predicting the primary diagnosis category for each patient admission, which helps in understanding patient health status and determining appropriate care pathways.
    
\textbf{Readmission Prediction:} We predict the likelihood of patient readmission within a specified timeframe after the discharge. This task helps in identifying patients at high risk of readmission, allowing healthcare providers to take proactive measures to reduce readmission rates and improve patient outcomes.

\textbf{Specific Diagnosis Readmission Prediction:} Beyond general readmission, we also focus on predicting readmissions for specific diagnoses such as Renal, Pulmonary, Infectious, Gastrointestinal, Oncology, and Neurological conditions. This targeted approach helps in understanding the nuances of each condition and tailoring patient care more effectively.


\subsection{Comparison with Baseline Models}
In our evaluation, we compare the performance of our proposed HealthGAT model against baseline classification methods. To establish a meaningful comparison, we employ logistic regression~\cite{lavalley2008logistic} as a baseline classification algorithm, which serves as a common and interpretable benchmark for binary classification tasks. Specifically, logistic regression is applied to various embeddings or representations of patient data, including \texttt{ME2Vec}, \texttt{metapath2vec}, \texttt{node2vec}, \texttt{LINE}, \texttt{Spectral Clustering}, and \texttt{NMF}.

By connecting our method to logistic regression and evaluating its performance alongside these models, we aim to assess the effectiveness and novelty of our proposed approach in predicting patient readmission.

\textbf{NMF (Non-negative Matrix Factorization)}: A traditional method that decomposes the dataset into non-negative matrices, often used for dimensionality reduction and feature extraction.

\textbf{SC (Spectral Clustering)}: An approach that utilizes eigenvalues of a similarity matrix to perform dimensionality reduction before clustering in fewer dimensions.

 \textbf{LINE (Large-scale Information Network Embedding)}~\cite{tang2015line}: Designed for embedding very large information networks, it preserves both local and global network structures.

\textbf{node2vec}~\cite{grover2016node2vec}: An algorithmic framework for learning continuous feature representations for nodes in networks, which efficiently leverages network neighborhood information.

 \textbf{metapath2vec}~\cite{dong2017metapath2vec}: Designed for heterogeneous networks, this method generates node embeddings by formalizing meta-path-based random walks to capture the semantic relationships between nodes.

 \textbf{ME2Vec}~\cite{wu2021leveraging}: A novel approach that combines multiple types of embeddings to capture various aspects of data in EHRs.

\subsection{Downstream Task Performance}

\subsubsection{Node Classification}
For the node classification task, we use the patient's medical journey, where we have their diagnosis history mentioning their service IDs. We use the patient embeddings and the diagnostic labels for training and predicting the diagnostic labels as our node classification task. In more detail, we first train the model on the training dataset (80\%) to obtain visit embeddings. Next, we use the visit embeddings in the training set (80\%) and their diagnosis labels to train a logistic regression classifier with L2 regularization. 
The model uses a logistic function for binary dependent variables, but it can also be extended to multi-class classification. This aligns well with categorizing patient records into multiple diagnosis categories~\cite{wu2010prediction}.

For evaluation, we employ Micro and Macro F1 scores. The Micro F1 Score offers insight into the overall effectiveness of the model across all predictions, treating each prediction equally regardless of class distribution. In contrast, the Macro F1 Score provides a class-level assessment of the model's performance, giving equal weight to each class and therefore emphasizing performance on minority classes. By considering both metrics, we gain a comprehensive understanding of the model's performance across all diagnostic labels.

As shown in Table \ref{tab:node_classification}, our model demonstrates proficiency in diagnosing various medical diagnoses. The model's performance across six different diagnosis labels demonstrates its understanding of complex medical data. Notably, in categories like Pulmonary and Gastrointestinal, the model achieves high scores in both Micro and Macro F1 scores, indicating accuracy and effective handling of class imbalances. However, in labels such as Infectious and Oncology, despite high Micro F1 scores, the Macro F1 scores are relatively lower. The overall Micro and Macro F1 scores, 0.926 and 0.529, respectively, underscore the model's general accuracy and capability to manage diverse diagnostic categories. The high Micro F1 score across all labels reflects the model's effectiveness in accurate predictions. In contrast, the Macro F1 score, though lower, is indicative of the challenges inherent in achieving balance across varied and unevenly represented classes.

Table \ref{tab:Baselines_classification} compares HealthGAT to baseline models in medical diagnostics. Our model, using a GAT with two auxiliary tasks, outperforms traditional models. HealthGAT achieves a Micro F1 score of 0.926 and a Macro F1 score of 0.529, significantly higher than baseline models like \texttt{ME2Vec}, \texttt{metapath2vec}, \texttt{node2vec}, and \texttt{LINE}. Incorporating advanced network architectures like GAT, which capture complex dependencies in data, and auxiliary tasks enhances the model's learning context and highlights the effectiveness of our approach.


\begin{table}[h]
\centering
\caption{Results of Node Classification Task}
\label{tab:node_classification}
\begin{tabular}{l c c c}
\toprule
Label Name & Micro F1 & Macro F1 \\ \hline
Renal               & 0.965            & 0.621         \\     
Pulmonary           & 0.955           & 0.954            \\   
Infectious          & 0.989           & 0.497          \\     
Gastrointestinal    & 0.983            & 0.969           \\   
Oncology            & 0.998           & 0.499           \\   
Neurologic          & 0.963           & 0.961           \\ \hline
Overall    & 0.926   & 0.529   \\ \bottomrule
\end{tabular}
\end{table}

\begin{table}[H]
  \caption{Comparison with Baselines for Classification Task}
  \label{tab:Baselines_classification}
  \centering
  \begin{tabular}{lcc}
    \toprule 
    Model&Micro-F1&Macro-F1\\  
    \midrule
    ME2Vec&0.879&0.676\\
    metapath2vec&0.870&0.577\\
    node2vec (service)&0.878&0.640\\
    node2vec (doctor)&0.861&0.463\\
    LINE (service)&0.866&0.586\\
    LINE (doctor)&0.861&0.463\\
    Spectral Clustering (service)&0.868&0.465\\
    Spectral Clustering (doctor)&0.868&0.465\\
    NMF (service)&0.879&0.600\\
    NMF (doctor)&0.867&0.469\\ \hline
    HealthGAT &0.926&0.529\\
    
    \bottomrule   
\end{tabular}
\end{table} 

\subsubsection{Readmission Prediction}
We evaluate our method using two different prediction tasks. First, we evaluate our method in readmission tasks and compare the performance with other baselines (Table \ref{tab:pred1}). Second, we evaluate the prediction task to predict readmission for specific diagnosis labels (Table \ref{tab:pred2}). We train a logistic regression classifier with visit embeddings and output labels. 

\begin{table}[H]
  \caption{ Performance Comparison with Baselines for Readmission Task}
  \label{tab:pred1}
  \centering
  \begin{tabular}{l cc}
  
    \toprule
    
    Model&AUROC&AUPRC\\  
    \midrule
    NMF&0.524&0.164\\
    SC&0.560&0.177\\
    LINE&0.570&0.178\\
    node2vec&0.573&0.181\\
    metapath2vec&0.576&0.183\\
    ME2Vec&0.588&0.186\\ \hline
    HealthGAT&0.59&0.20\\
    \bottomrule

\end{tabular}

\end{table} 

\begin{table}[H]
  \caption{Readmission for each diagnosis separately}
  \label{tab:pred2}
  \centering
  \begin{tabular}{ccccc}
  
    \toprule
    
    Prevalance&Model&AUROC&AUPRC&F1\\  
    \midrule
    58\%&Renal admission&0.57&0.46&0.58\\
    55\%&Pulmonary admission&0.56&0.43&0.60\\
    18\%&Infectious admission &0.67&0.24&0.84\\
    17\%&Gastrointestinal admission&0.62&0.20&0.85\\
    7\%&Ocncology admission&0.85&0.31&0.92\\
    33\%&Neurologic admission&0.63&0.3&0.74\\
    8.4\%&Mortaity Prediction&0.70&0.16&0.91\\
    \bottomrule       
\end{tabular}
\end{table} 

The performance of several models, including HealthGAT, in predicting readmissions is compared in Table \ref{tab:pred1}. Area Under the Receiver Operating Characteristic curve (AUROC) and Area Under the Precision-Recall curve (AUPRC) are the measures utilized for evaluation. Higher AUPRC and AUROC scores signify improved prediction performance. When compared to more established techniques like \texttt{NMF}, \texttt{SC}, \texttt{LINE}, \texttt{node2vec}, \texttt{metapath2vec}, and \texttt{ME2Vec}, HealthGAT has the best AUROC of 0.59 and AUPRC of 0.20 among the baseline models, indicating its efficacy in readmission prediction.

The outcomes of readmission prediction for particular diagnosis labels are shown in Table \ref{tab:pred2}. A distinct diagnosis category and the dataset's prevalence for it are indicated by each row. The table presents the AUROC, AUPRC, and F1 score for each diagnosis. HealthGAT is notable for achieving competitive performance in a number of diagnosis categories. For cancer admission and mortality prediction, particularly high AUROC and F1 scores are noted, indicating the model's capacity to precisely forecast readmissions for these types of diseases.

\section{Future Directions and Enhancements}
As we refine and advance our HealthGAT model, we will explore a range of novel strategies and assessments to elevate its performance in processing EHRs.

\subsection{Strategic Developments}
One next step would involve broadening the scope of medical code embeddings by including a wider array of EHR components such as vital signs, medication records, and procedural data. This expansion aims to deepen the model's understanding of patient health narratives. Another step relates to investigating the role of pre-training tasks in improving the process of deriving patient embeddings from visit data. These new tasks can be designed to predict different elements of future medical records, such as upcoming visit details and masked medication information. This approach will not only enrich the embeddings but also ensure our model's adaptability to various EHR scenarios.

\subsection{ Evaluation Techniques}
We plan to evaluate our model using different time segments. Currently, a 1-hour window is used. Experimenting with shorter or longer time windows could reveal how the temporal proximity of medical services affects their co-occurrence and, consequently, the structure of the graph. For instance, a shorter window might capture more immediate service relationships, while a longer window could reveal broader treatment patterns. Our evaluation efforts will prioritize assessing other advanced models combined with GNN, particularly those based on sequential learning and transformer architectures, for deriving patient embeddings from visit data. This exploration will leverage state-of-the-art techniques to maximize the potential of visit embeddings generated by our approach.

\subsection{Limitations and Potential Biases}
We acknowledge that the HealthGAT model, like any machine learning model, may have inherent limitations and potential biases. These could stem from various factors such as data quality, representativeness of the training data, missing data, or the algorithm's design and implementation. Future work will involve conducting a thorough analysis to identify and address these limitations and biases \cite{poulainfacct}. This will include assessing the model's performance across diverse demographic groups, clinical settings, and types of medical conditions to ensure its generalizability and fairness. Additionally, we will explore techniques for mitigating biases in both the data and the model itself, such as through data augmentation, fairness-aware training, and interpretability measures. By proactively addressing these challenges, we aim to enhance the reliability and equity of the HealthGAT model in real-world healthcare applications.

\section{Conclusion}
Our research used GNNs to extract graphical structures from EHR entities. We pre-trained EHR-specific data representations inspired by pre-training in language models. Pre-trained embeddings facilitate the development of predictive models and their transferability allows for versatile application across tasks. We enriched EHR data by incorporating auxiliary tasks and exploiting hierarchical knowledge, leading to better performance and understanding of medical conditions. Our research developed a GNN-based model, HealthGAT, to capture intricate correlations in medical data for accurate predictions in healthcare. We conducted node classification tasks for various diagnostic categories, including general readmission prediction and specific predictions for patient readmissions related to Renal, Pulmonary, Infectious, Gastrointestinal, Oncology, and Neurological conditions. Our approach outperformed established baseline models in node classification and readmission prediction tasks, as evidenced by higher AUROC, AUPRC, and F1 scores. Our GNN-based approach for node classification in EHR data can mark a stride towards enhancing patient care. Our method offers improved prediction accuracy and overcomes challenges posed by complex disease characteristics and data scarcity. The insights gained from our experiments pave the way for future advancements in EHR data analysis and patient care optimization.

\section{Acknowledgements} Our study was supported by NIH awards, P20GM103446 and P20GM113125.

\bibliographystyle{IEEEtran}
\bibliography{Camera_Ready_Version_HealthGAT.bib}

\begin{thebibliography}{10}
\providecommand{\url}[1]{#1}
\csname url@samestyle\endcsname
\providecommand{\newblock}{\relax}
\providecommand{\bibinfo}[2]{#2}
\providecommand{\BIBentrySTDinterwordspacing}{\spaceskip=0pt\relax}
\providecommand{\BIBentryALTinterwordstretchfactor}{4}
\providecommand{\BIBentryALTinterwordspacing}{\spaceskip=\fontdimen2\font plus
\BIBentryALTinterwordstretchfactor\fontdimen3\font minus \fontdimen4\font\relax}
\providecommand{\BIBforeignlanguage}[2]{{%
\expandafter\ifx\csname l@#1\endcsname\relax
\typeout{** WARNING: IEEEtran.bst: No hyphenation pattern has been}%
\typeout{** loaded for the language `#1'. Using the pattern for}%
\typeout{** the default language instead.}%
\else
\language=\csname l@#1\endcsname
\fi
#2}}
\providecommand{\BIBdecl}{\relax}
\BIBdecl

\bibitem{golmaei2021deepnote}
S.~N. Golmaei and X.~Luo, ``Deepnote-gnn: predicting hospital readmission using clinical notes and patient network,'' in \emph{Proceedings of the 12th ACM Conference on Bioinformatics, Computational Biology, and Health Informatics}, 2021, pp. 1--9.

\bibitem{4700287}
F.~Scarselli, M.~Gori, A.~C. Tsoi, M.~Hagenbuchner, and G.~Monfardini, ``The graph neural network model,'' \emph{IEEE Transactions on Neural Networks}, vol.~20, no.~1, pp. 61--80, 2009.

\bibitem{jg2022graph}
D.~O. JG and F.~E. Mustafa, ``Graph neural network modelling as a potentially effective method for predicting and analyzing procedures based on patients' diagnoses.'' \emph{Artificial Intelligence in Medicine}, vol. 131, pp. 102\,359--102\,359, 2022.

\bibitem{wanyan2021deep}
T.~Wanyan, H.~Honarvar, A.~Azad, Y.~Ding, and B.~S. Glicksberg, ``Deep learning with heterogeneous graph embeddings for mortality prediction from electronic health records,'' \emph{Data Intelligence}, vol.~3, no.~3, pp. 329--339, 2021.

\bibitem{liu2022introduction}
Z.~Liu and J.~Zhou, \emph{Introduction to graph neural networks}.\hskip 1em plus 0.5em minus 0.4em\relax Springer Nature, 2022.

\bibitem{atwood2016diffusion}
J.~Atwood and D.~Towsley, ``Diffusion-convolutional neural networks,'' \emph{Advances in neural information processing systems}, vol.~29, 2016.

\bibitem{li2018deeper}
Q.~Li, Z.~Han, and X.-M. Wu, ``Deeper insights into graph convolutional networks for semi-supervised learning,'' in \emph{Proceedings of the AAAI conference on artificial intelligence}, vol.~32, no.~1, 2018.

\bibitem{kazienko2012label}
P.~Kazienko and T.~Kajdanowicz, ``Label-dependent node classification in the network,'' \emph{Neurocomputing}, vol.~75, no.~1, pp. 199--209, 2012.

\bibitem{quinn2019electronic}
M.~Quinn, J.~Forman, M.~Harrod, S.~Winter, K.~E. Fowler, S.~L. Krein, A.~Gupta, S.~Saint, H.~Singh, and V.~Chopra, ``Electronic health records, communication, and data sharing: challenges and opportunities for improving the diagnostic process,'' \emph{Diagnosis}, vol.~6, no.~3, pp. 241--248, 2019.

\bibitem{manessi2021graph}
F.~Manessi and A.~Rozza, ``Graph-based neural network models with multiple self-supervised auxiliary tasks,'' \emph{Pattern Recognition Letters}, vol. 148, pp. 15--21, 2021.

\bibitem{NEURIPS2018_934b5358}
\BIBentryALTinterwordspacing
E.~Choi, C.~Xiao, W.~Stewart, and J.~Sun, ``Mime: Multilevel medical embedding of electronic health records for predictive healthcare,'' in \emph{Advances in Neural Information Processing Systems}, S.~Bengio, H.~Wallach, H.~Larochelle, K.~Grauman, N.~Cesa-Bianchi, and R.~Garnett, Eds., vol.~31.\hskip 1em plus 0.5em minus 0.4em\relax Curran Associates, Inc., 2018. [Online]. Available: \url{https://proceedings.neurips.cc/paper_files/paper/2018/file/934b535800b1cba8f96a5d72f72f1611-Paper.pdf}
\BIBentrySTDinterwordspacing

\bibitem{grover2016node2vec}
A.~Grover and J.~Leskovec, ``node2vec: Scalable feature learning for networks,'' in \emph{Proceedings of the 22nd ACM SIGKDD international conference on Knowledge discovery and data mining}, 2016, pp. 855--864.

\bibitem{velickovic2017graph}
P.~Velickovic, G.~Cucurull, A.~Casanova, A.~Romero, P.~Lio, Y.~Bengio \emph{et~al.}, ``Graph attention networks,'' \emph{stat}, vol. 1050, no.~20, pp. 10--48\,550, 2017.

\bibitem{pollard2018eicu}
T.~J. Pollard, A.~E. Johnson, J.~D. Raffa, L.~A. Celi, R.~G. Mark, and O.~Badawi, ``The eicu collaborative research database, a freely available multi-center database for critical care research,'' \emph{Scientific data}, vol.~5, no.~1, pp. 1--13, 2018.

\bibitem{guptamimic2022}
\BIBentryALTinterwordspacing
M.~Gupta, B.~Gallamoza, N.~Cutrona, P.~Dhakal, R.~Poulain, and R.~Beheshti, ``An extensive data processing pipeline for mimic-iv,'' in \emph{Proceedings of the 2nd Machine Learning for Health symposium}, ser. Proceedings of Machine Learning Research, A.~Parziale, M.~Agrawal, S.~Joshi, I.~Y. Chen, S.~Tang, L.~Oala, and A.~Subbaswamy, Eds., vol. 193.\hskip 1em plus 0.5em minus 0.4em\relax PMLR, 28 Nov 2022, pp. 311--325. [Online]. Available: \url{https://proceedings.mlr.press/v193/gupta22a.html}
\BIBentrySTDinterwordspacing

\bibitem{GuptaMedRxiv}
\BIBentryALTinterwordspacing
M.~Gupta, T.-L.~T. Phan, D.~Eckrich, H.~T. Bunnell, and R.~Beheshti, ``Reliable prediction of childhood obesity using only routinely collected ehrs is possible,'' \emph{medRxiv}, 2024. [Online]. Available: \url{https://www.medrxiv.org/content/early/2024/01/31/2024.01.29.24301945}
\BIBentrySTDinterwordspacing

\bibitem{sun2020disease}
Z.~Sun, H.~Yin, H.~Chen, T.~Chen, L.~Cui, and F.~Yang, ``Disease prediction via graph neural networks,'' \emph{IEEE Journal of Biomedical and Health Informatics}, vol.~25, no.~3, pp. 818--826, 2020.

\bibitem{landi2020deep}
I.~Landi, B.~S. Glicksberg, H.-C. Lee, S.~Cherng, G.~Landi, M.~Danieletto, J.~T. Dudley, C.~Furlanello, and R.~Miotto, ``Deep representation learning of electronic health records to unlock patient stratification at scale,'' \emph{NPJ digital medicine}, vol.~3, no.~1, p.~96, 2020.

\bibitem{li2020graph}
Y.~Li, B.~Qian, X.~Zhang, and H.~Liu, ``Graph neural network-based diagnosis prediction,'' \emph{Big Data}, vol.~8, no.~5, pp. 379--390, 2020.

\bibitem{poulain2024graph}
\BIBentryALTinterwordspacing
R.~Poulain and R.~Beheshti, ``Graph transformers on {EHR}s: Better representation improves downstream performance,'' in \emph{The Twelfth International Conference on Learning Representations}, 2024. [Online]. Available: \url{https://openreview.net/forum?id=pe0Vdv7rsL}
\BIBentrySTDinterwordspacing

\bibitem{perozzi2014deepwalk}
B.~Perozzi, R.~Al-Rfou, and S.~Skiena, ``Deepwalk: Online learning of social representations,'' in \emph{Proceedings of the 20th ACM SIGKDD international conference on Knowledge discovery and data mining}, 2014, pp. 701--710.

\bibitem{cao2015grarep}
S.~Cao, W.~Lu, and Q.~Xu, ``Grarep: Learning graph representations with global structural information,'' in \emph{Proceedings of the 24th ACM international on conference on information and knowledge management}, 2015, pp. 891--900.

\bibitem{ou2016asymmetric}
M.~Ou, P.~Cui, J.~Pei, Z.~Zhang, and W.~Zhu, ``Asymmetric transitivity preserving graph embedding,'' in \emph{Proceedings of the 22nd ACM SIGKDD international conference on Knowledge discovery and data mining}, 2016, pp. 1105--1114.

\bibitem{tang2015line}
J.~Tang, M.~Qu, M.~Wang, M.~Zhang, J.~Yan, and Q.~Mei, ``Line: Large-scale information network embedding,'' in \emph{Proceedings of the 24th international conference on world wide web}, 2015, pp. 1067--1077.

\bibitem{hamilton2017inductive}
W.~Hamilton, Z.~Ying, and J.~Leskovec, ``Inductive representation learning on large graphs,'' \emph{Advances in neural information processing systems}, vol.~30, 2017.

\bibitem{jiang2020maru}
J.-Y. Jiang, Z.~Li, C.~J.-T. Ju, and W.~Wang, ``Maru: Meta-context aware random walks for heterogeneous network representation learning,'' in \emph{Proceedings of the 29th ACM International Conference on Information \& Knowledge Management}, 2020, pp. 575--584.

\bibitem{dong2017metapath2vec}
Y.~Dong, N.~V. Chawla, and A.~Swami, ``metapath2vec: Scalable representation learning for heterogeneous networks,'' in \emph{Proceedings of the 23rd ACM SIGKDD international conference on knowledge discovery and data mining}, 2017, pp. 135--144.

\bibitem{chang2015heterogeneous}
S.~Chang, W.~Han, J.~Tang, G.-J. Qi, C.~C. Aggarwal, and T.~S. Huang, ``Heterogeneous network embedding via deep architectures,'' in \emph{Proceedings of the 21th ACM SIGKDD international conference on knowledge discovery and data mining}, 2015, pp. 119--128.

\bibitem{wu2021leveraging}
T.~Wu, Y.~Wang, Y.~Wang, E.~Zhao, and Y.~Yuan, ``Leveraging graph-based hierarchical medical entity embedding for healthcare applications,'' \emph{Scientific reports}, vol.~11, no.~1, p. 5858, 2021.

\bibitem{talati2021deep}
N.~Talati, D.~Jin, H.~Ye, A.~Brahmakshatriya, G.~Dasika, S.~Amarasinghe, T.~Mudge, D.~Koutra, and R.~Dreslinski, ``A deep dive into understanding the random walk-based temporal graph learning,'' in \emph{2021 IEEE International Symposium on Workload Characterization (IISWC)}.\hskip 1em plus 0.5em minus 0.4em\relax IEEE, 2021, pp. 87--100.

\bibitem{lu2023disease}
H.~Lu and S.~Uddin, ``Disease prediction using graph machine learning based on electronic health data: A review of approaches and trends,'' in \emph{Healthcare}, vol.~11, no.~7.\hskip 1em plus 0.5em minus 0.4em\relax MDPI, 2023, p. 1031.

\bibitem{mikolov2013distributed}
T.~Mikolov, I.~Sutskever, K.~Chen, G.~S. Corrado, and J.~Dean, ``Distributed representations of words and phrases and their compositionality,'' \emph{Advances in neural information processing systems}, vol.~26, 2013.

\bibitem{gao2022mtgnn}
J.~Gao, X.~Zhang, L.~Tian, Y.~Liu, J.~Wang, Z.~Li, and X.~Hu, ``Mtgnn: multi-task graph neural network based few-shot learning for disease similarity measurement,'' \emph{Methods}, vol. 198, pp. 88--95, 2022.

\bibitem{choi2017gram}
E.~Choi, M.~T. Bahadori, L.~Song, W.~F. Stewart, and J.~Sun, ``Gram: graph-based attention model for healthcare representation learning,'' in \emph{Proceedings of the 23rd ACM SIGKDD international conference on knowledge discovery and data mining}, 2017, pp. 787--795.

\bibitem{wu2021enhancing}
Y.~Wu, Y.~Song, H.~Huang, F.~Ye, X.~Xie, and H.~Jin, ``Enhancing graph neural networks via auxiliary training for semi-supervised node classification,'' \emph{Knowledge-Based Systems}, vol. 220, p. 106884, 2021.

\bibitem{lv2023semi}
J.~Lv, K.~Song, Q.~Ye, and G.~Tian, ``Semi-supervised node classification via fine-grained graph auxiliary augmentation learning,'' \emph{Pattern Recognition}, vol. 137, p. 109301, 2023.

\bibitem{lavalley2008logistic}
M.~P. LaValley, ``Logistic regression,'' \emph{Circulation}, vol. 117, no.~18, pp. 2395--2399, 2008.

\bibitem{wu2010prediction}
J.~Wu, J.~Roy, and W.~F. Stewart, ``Prediction modeling using ehr data: challenges, strategies, and a comparison of machine learning approaches,'' \emph{Medical care}, pp. S106--S113, 2010.

\bibitem{poulainfacct}
\BIBentryALTinterwordspacing
R.~Poulain, M.~F. Bin~Tarek, and R.~Beheshti, ``Improving fairness in ai models on electronic health records: The case for federated learning methods,'' in \emph{Proceedings of the 2023 ACM Conference on Fairness, Accountability, and Transparency}, ser. FAccT '23.\hskip 1em plus 0.5em minus 0.4em\relax New York, NY, USA: Association for Computing Machinery, 2023, p. 1599–1608. [Online]. Available: \url{https://doi.org/10.1145/3593013.3594102}
\BIBentrySTDinterwordspacing

\end{thebibliography}
\end{document}